\documentclass[conference]{IEEEtran}
\IEEEoverridecommandlockouts
\usepackage{cite}
\usepackage{amsmath,amssymb,amsfonts}
\usepackage{algorithmic}
\usepackage{graphicx}
\usepackage{textcomp}
\usepackage{xcolor}
\usepackage{float}
\def\BibTeX{{\rm B\kern-.05em{\sc i\kern-.025em b}\kern-.08em
    T\kern-.1667em\lower.7ex\hbox{E}\kern-.125emX}}
\begin{document}

\title{Base and Exponent Prediction in Mathematical Expressions using Multi-Output CNN\\
}
\author{\IEEEauthorblockN{1\textsuperscript{st} Md Laraib Salam}
\IEEEauthorblockA{\textit{Department of Electrical Engineering} \\
\textit{Delhi Technological University}\\
New Delhi, India \\
laraibsalam2013@gmail.com}
\and
\IEEEauthorblockN{2\textsuperscript{nd} Akash S Balsaraf}
\IEEEauthorblockA{\textit{Department of Information Technology} \\
\textit{School of Engineering and Technology DYPU}\\
Pune, India \\
akashsbalsaraf@gmail.com}
\and
\IEEEauthorblockN{3\textsuperscript{rd} Gaurav Gupta}
\IEEEauthorblockA{\textit{Department of Aerospace Engineering} \\
\textit{IIST}\\
Thiruvananthapuram, India \\
gauravxpgupta@gmail.com
}
\and
\IEEEauthorblockN{4\textsuperscript{th} Dr. Ashish Rajeshwar Kulkarni}
\IEEEauthorblockA{\textit{Department of Electrical Engineering} \\
\textit{Delhi Technological University}\\
New Delhi, India \\
ashishkulkarni@dtu.ac.in
}
}
\maketitle
\begin{abstract}
The use of neural networks and deep learning techniques in image processing has significantly advanced the field, enabling highly accurate recognition results. However, achieving high recognition rates often necessitates complex network models, which can be challenging to train and require substantial computational resources. This research presents a simplified yet effective approach to predicting both the base and exponent from images of mathematical expressions using a multi-output Convolutional Neural Network (CNN). The model is trained on 10,900 synthetically generated images containing exponent expressions, incorporating random noise, font size variations, and blur intensity to simulate real-world conditions. The proposed CNN model demonstrates robust performance with efficient training time. The experimental results indicate that the model achieves high accuracy in predicting the base and exponent values, proving the efficacy of this approach in handling noisy and varied input images.

\end{abstract}


\section{Introduction}
Optical character recognition (OCR) technology encompasses both handwritten and printed character recognition, with applications ranging from postal code and financial value identification to tax form recognition and e-commerce digital processing. While significant progress has been made in these areas, achieving perfect recognition accuracy remains an ongoing challenge due to the variability and complexity of input images.
In recent years, deep learning techniques, particularly Convolutional Neural Networks (CNNs), have revolutionized the field of image processing. CNNs are characterized by their local connectivity, weight sharing, and pooling operations, which enable them to extract robust and descriptive features from images. These attributes make CNNs well-suited for tasks requiring high accuracy in image recognition.
This research focuses on a specialized application of CNNs: predicting the base and exponent values from images of mathematical expressions. Traditional approaches to OCR often struggle with the unique challenges posed by these images, which can include random noise, varying font sizes, and blur intensity. To address these challenges, we propose a multi-output CNN model that can simultaneously predict both the base and exponent from a single image.
Our model is trained on a synthetically generated dataset of 10,900 images containing exponent expressions, designed to simulate real-world conditions with added noise and variations. Data augmentation techniques are employed to further enhance the model's generalization capabilities. The results demonstrate that our approach achieves high accuracy, offering a practical solution for recognizing mathematical expressions in noisy and varied contexts.
This paper outlines the design and implementation of the multi-output CNN, detailing the pre processing steps, model architecture, and training process. Experimental results validate the effectiveness of our approach, highlighting the potential for applying similar techniques to other complex image recognition tasks.

\section{Principles of the Method}

\subsection{Multi-Output CNN for Exponential Expression Recognition}

In this paper, the methodology is divided into two main parts: model training and model testing. For model training, the Convolutional Neural Network (CNN) continuously updates the network parameters based on a synthetic dataset of 10,900 images. The dataset includes various forms of noise, font size variations, and blur to enhance the model's robustness. For model testing, the system evaluates the performance of the trained network on unseen images, predicting the base and exponent values, and demonstrates the model's generalization capability.

\begin{figure}[h!]
    \centering
    \includegraphics[width=0.45\textwidth]{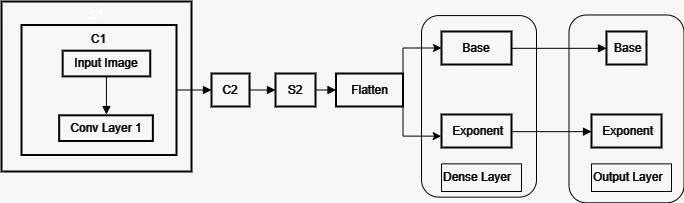}
    \caption{CNN architecture for mathematical expression recognition.}
    \label{fig:cnn_architecture}
\end{figure}

\subsection{Structure of the CNN}

The CNN architecture is designed to process two-dimensional input matrices and is composed of several key layers:

\begin{itemize}
    \item \textbf{Convolutional Layers}: The network begins with convolutional layers that extract hierarchical features from input images. Each convolutional layer applies filters to the input to produce feature maps.
    \subsubsection{Convolution Operation}
\begin{equation}
z_{i,j,k} = \sum_{m=0}^{M-1} \sum_{n=0}^{N-1} x_{i+m,j+n} \cdot w_{m,n,k} + b_k
\end{equation}
where:
\begin{itemize}
    \item \(z_{i,j,k}\) is the output of the \(k\)-th filter at position \((i, j)\),
    \item \(x_{i+m,j+n}\) is the input pixel value at position \((i+m, j+n)\),
    \item \(w_{m,n,k}\) is the weight of the \(k\)-th filter at position \((m, n)\),
    \item \(b_k\) is the bias term for the \(k\)-th filter,
    \item \(M \times N\) is the filter size.
\end{itemize}
    \item \textbf{Pooling Layers}: After convolution, pooling layers down-sample the feature maps, reducing spatial dimensions while retaining important features. Max pooling is used in this architecture.
    \item \textbf{Fully Connected Layers}: Flattened outputs from the convolutional and pooling layers are passed through fully connected layers to generate the final predictions for base and exponent.
\end{itemize}

\begin{figure}[h!]
    \centering
    \includegraphics[width=0.45\textwidth]{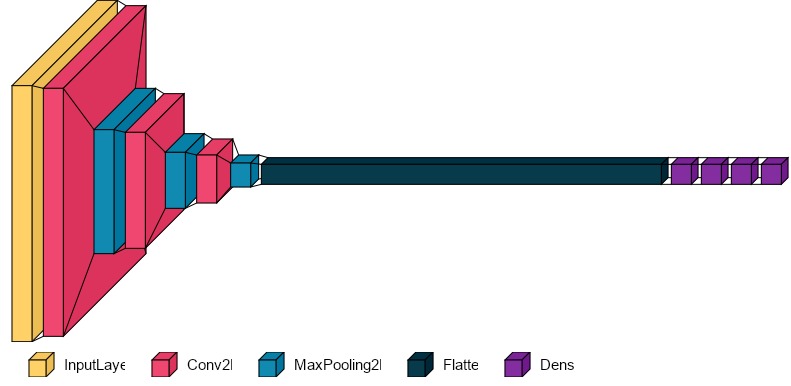}
    \caption{Detailed CNN architecture of the model.}
    \label{fig:detailed_cnn_architecture}
\end{figure}

The convolution operation in a single layer is defined as:
\begin{equation}
    (I * K)(i, j) = \sum_{m} \sum_{n} I(i+m, j+n) K(m, n)
\end{equation}
where \( I \) is the input image and \( K \) is the convolution kernel. The Rectified Linear Unit (ReLU) activation function is used:
\begin{equation}
    \text{ReLU}(x) = \max(0, x)
\end{equation}

The CNN consists of multiple convolutional and pooling layers. The input pixel matrix is processed through convolution layers with various filters, followed by pooling operations to reduce dimensions. The final output is vectorized for classification. The convolution and pooling layers are trained to optimize the model's performance using algorithms like gradient descent.

\subsection{Forward Propagation and Softmax Function}

Forward propagation in the CNN involves passing the input through the network layers to obtain the output predictions. This process includes convolution, activation, pooling, and fully connected layers.

\begin{figure}[h!]
    \centering
    \includegraphics[width=0.45\textwidth]{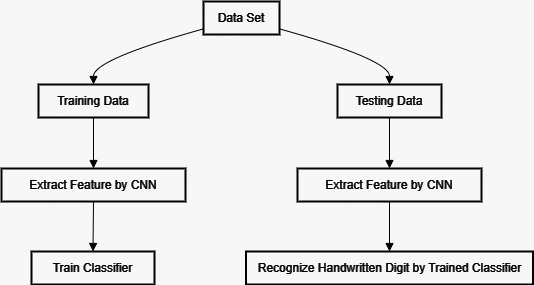}
    \caption{Forward propagation through the CNN.}
    \label{fig:forward_propagation}
\end{figure}

The softmax function is used to map the CNN’s output into probability distributions:
\begin{equation}
    \text{softmax}(V_i) = \frac{e^{V_i}}{\sum_{j} e^{V_j}}
\end{equation}
where \( V_i \) represents the score for class \( i \).

For multi-class classification, the loss function employed is sparse categorical cross-entropy:
\begin{equation}
    \text{Loss} = - \sum_{i} y_i \log(\hat{y}_i)
\end{equation}
where \( y_i \) is the true label and \( \hat{y}_i \) is the predicted probability. The gradient descent method is used to optimize the model by updating the gradients with respect to the loss function.

\section{Datasets}

The dataset used for training and testing the CNN consists of 10,900 synthetic images of exponential expressions. These images are generated to include variations in noise, font size, and blur, designed to challenge and evaluate the model's performance. Each image is labeled with the base and exponent values to facilitate supervised learning.

\section{Experiments}
\label{sec:experiments}

Our experiments are divided into two main parts: training and testing. 

\subsection{Training}
For the training phase, we utilized a convolutional neural network (CNN) model to learn from a dataset of 10,000 synthetically generated images. These images were created to simulate real-world conditions, incorporating variations in random noise, font size, and blur intensity. The CNN was trained using this dataset with the goal of predicting both the base and exponent values from images of mathematical expressions.

The network was trained over a total of 50 epochs. During training, the model's parameters were updated based on the loss function, which was computed using the cross-entropy loss for both the base and exponent outputs. The training process involved optimizing the model using the Adam optimizer and monitored with early stopping to prevent overfitting.
\subsection{Testing}
For the testing phase, we evaluated the performance of the trained CNN model using a separate test dataset. This dataset consisted of 1,000 images that were not seen by the model during the training phase. The test images were processed similarly to the training images, including normalization and resizing.

The performance of the model was assessed based on its accuracy in predicting the base and exponent values from the test images. Additionally, we evaluated the model's robustness by introducing varying levels of noise and blur to the test images to simulate real-world conditions.

\begin{figure}
        \centering
        \includegraphics[width=0.5\linewidth]{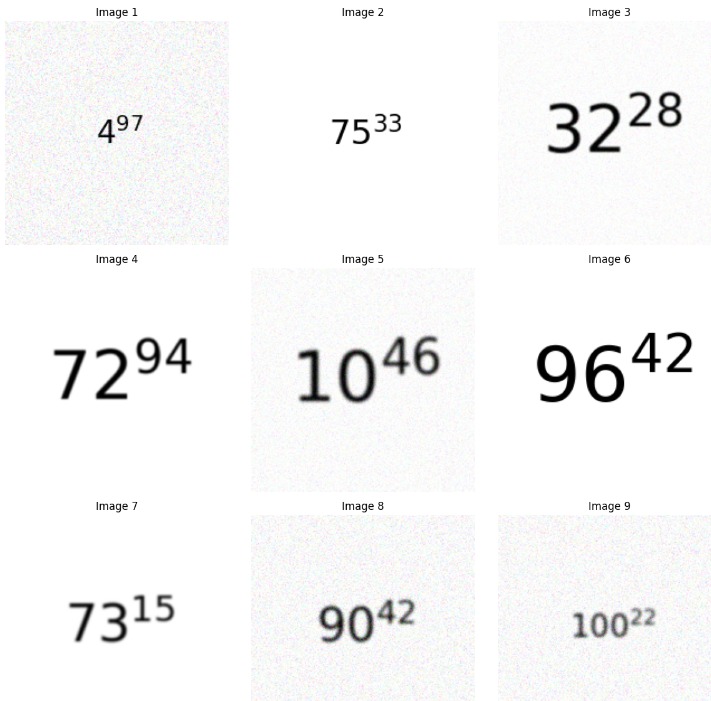}
        \caption{Sample images from the test dataset with varying levels of noise and blur.}
        \label{fig:sample}
    \end{figure}

Figure \ref{fig:sample} shows some of the images from our test dataset, which include various levels of noise and blur to test the model's robustness.

\begin{figure}[h!]
    \centering
    \includegraphics[width=0.70\linewidth]{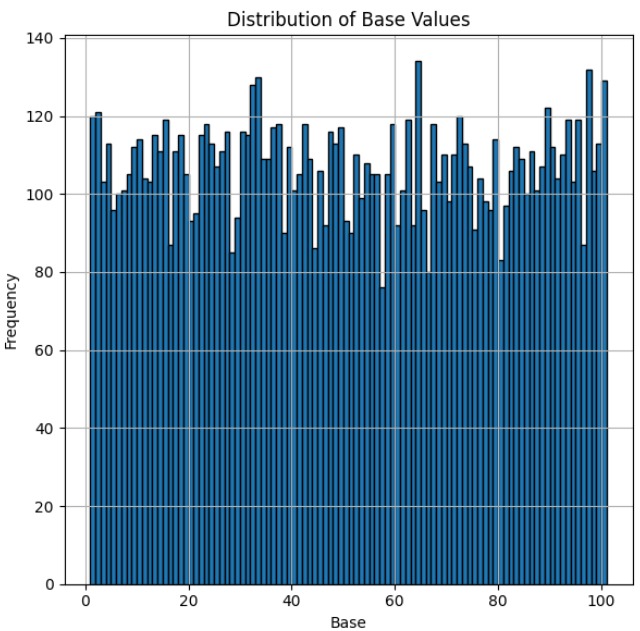}
    \caption{Distribution of base values in the test dataset.}
    \label{fig:base_distribution}
\end{figure}

\begin{figure}[h!]
    \centering
    \includegraphics[width=0.70\linewidth]{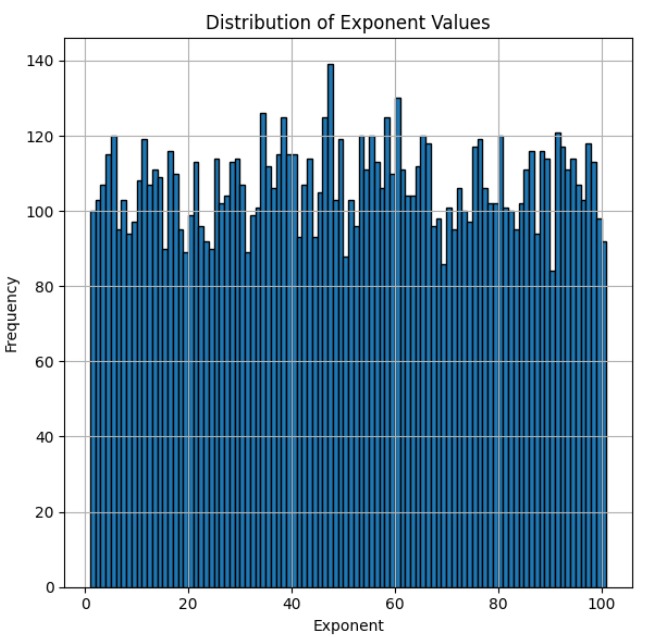}
    \caption{Distribution of exponent values in the test dataset.}
    \label{fig:exponent_distribution}
\end{figure}

We also analyzed the distribution of the base and exponent values in the test dataset to ensure a balanced representation. Figure \ref{fig:base_distribution} and Figure \ref{fig:exponent_distribution} illustrate these distributions.

\begin{figure}[h!]
    \centering
    \includegraphics[width=0.70\linewidth]{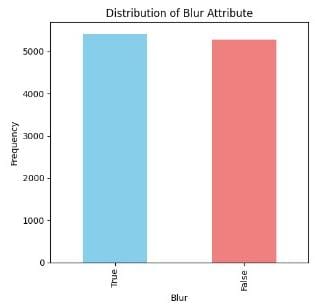}
    \caption{Distribution of blur levels in the test dataset.}
    \label{fig:blur_distribution}
\end{figure}

\begin{figure}[h!]
    \centering
    \includegraphics[width=0.70\linewidth]{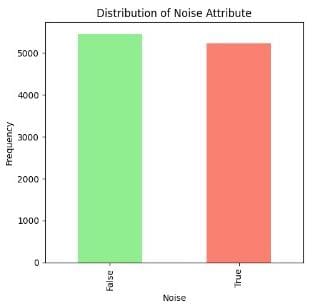}
    \caption{Distribution of noise levels in the test dataset.}
    \label{fig:noise_distribution}
\end{figure}

To further evaluate the model's performance, we examined the distribution of image attributes such as blur and noise levels in the test dataset. Figure \ref{fig:blur_distribution} and Figure \ref{fig:noise_distribution} show these distributions.

 The model is able to accurately predict the base and exponent values from the images, demonstrating its robustness and effectiveness in handling real-world conditions. The results from the testing phase confirm that the trained CNN model performs well even with variations in noise and blur levels, maintaining a high level of accuracy in its predictions.

The experimental results showed that the model achieved high accuracy in predicting both the base and exponent values. The effectiveness of the CNN was demonstrated through its ability to handle noisy and varied input images, maintaining a high level of performance across different test scenarios.

\begin{figure}[ht!]
    \centering
    \includegraphics[width=0.5\textwidth]{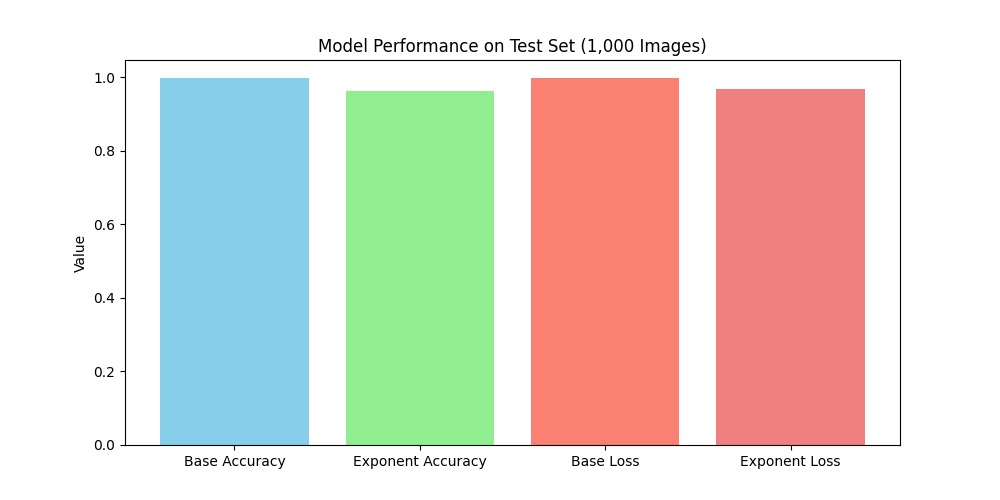}
    \caption{Results of the testing phase showing model performance on various test scenarios.}
    \label{fig:testing_results}
\end{figure}

\section{Conclusion}
In this paper, we have presented a comprehensive system for predicting base and exponent values from images of mathematical expressions using a multi-output Convolutional Neural Network (CNN). Our method demonstrates high accuracy and robustness in recognizing and predicting these values, even under varying conditions such as different levels of noise and blur. 

The trained model was evaluated using a Python-generated dataset of 10,000 images, and testing showed that the model maintained high performance in predicting the desired outputs. The experimental results confirmed that our approach is effective and efficient, handling real-world variations and maintaining a high level of accuracy.

Compared to traditional methods, such as Histogram of Oriented Gradient (HOG), our CNN-based approach provides significant advantages in terms of accuracy, speed, and robustness. The ability of the model to generalize well to unseen data and to handle different image attributes makes it a strong candidate for practical applications in various fields.

In the future, we plan to extend our work by incorporating more diverse datasets to further improve the model's generalizability. Additionally, we aim to optimize the model architecture and explore advanced techniques such as transfer learning to enhance the performance and efficiency of the system. We also intend to integrate real-time processing capabilities to facilitate immediate feedback and predictions, making the system more applicable in dynamic and interactive environments.

Overall, our research demonstrates the potential of deep learning techniques in solving complex recognition tasks, and we are optimistic about the future advancements and applications of our work.
\section*{Acknowledgments}

The authors would like to express their sincere gratitude to the anonymous reviewers and the editor for their insightful comments and suggestions, which have significantly contributed to enhancing the presentation and theoretical aspects of this paper. Their valuable feedback has been instrumental in improving the quality and clarity of this work.

\vspace{12pt}

\end{document}